\documentclass{article}

\usepackage{subcaption}
\usepackage{xcolor}
\usepackage{PRIMEarxiv}

\usepackage[utf8]{inputenc} % allow utf-8 input
\usepackage[T1]{fontenc}    % use 8-bit T1 fonts
\usepackage{hyperref}       % hyperlinks
\usepackage{url}            % simple URL typesetting
\usepackage{booktabs}       % professional-quality tables
\usepackage{amsfonts}       % blackboard math symbols
\usepackage{nicefrac}       % compact symbols for 1/2, etc.
\usepackage{microtype}      % microtypography
\usepackage{lipsum}
\usepackage{graphicx}
\graphicspath{{media/}}     % organize your images and other figures under media/ folder
\usepackage{amsmath}
\makeatletter
\newcommand{\ostar}{\mathbin{\mathpalette\make@circled\star}}
\newcommand{\make@circled}[2]{%
  \ooalign{$\m@th#1\smallbigcirc{#1}$\cr\hidewidth$\m@th#1#2$\hidewidth\cr}%
}
\newcommand{\smallbigcirc}[1]{%
  \vcenter{\hbox{\scalebox{0.77778}{$\m@th#1\bigcirc$}}}%
}
\makeatother
  
\title{Weakly Supervised Human Skin Segmentation using Guidance Attention Mechanisms
}

\author{
  Kooshan Hashemifard, Pau Climent-Perez, Francisco Florez-Revuelta \\
  Department of Computing Technology \\
  University of Alicante \\
  San Vicente Del Raspeig\\
  \texttt{\{k.hashemifard, pau.climent,francisco.florez\}@ua.es} \\
}

\begin{document}
\maketitle

\begin{abstract}
Human skin segmentation is a crucial task in computer vision and biometric systems, yet it poses several challenges such as variability in skin color, pose, and illumination. This paper presents a robust data-driven skin segmentation method for a single image that addresses these challenges through the integration of contextual information and efficient network design. In addition to robustness and accuracy, the integration into real-time systems requires a careful balance between computational power, speed, and performance. 
The proposed method incorporates two attention modules, Body Attention and Skin Attention, that utilize contextual information to improve segmentation results. These modules draw attention to the desired areas, focusing on the body boundaries and skin pixels, respectively. Additionally, an efficient network architecture is employed in the encoder part to minimize computational power while retaining high performance.
To handle the issue of noisy labels in skin datasets, the proposed method uses a weakly supervised training strategy, relying on the Skin Attention module. The results of this study demonstrate that the proposed method is comparable to, or outperforms, state-of-the-art methods on benchmark datasets.
\end{abstract}

% keywords can be removed
\keywords{Skin segmentation \and Attention mechanism \and Deep neural networks}

\section{Introduction}
Skin detection, also known as skin segmentation, is the process of separating skin pixels or regions in an image from non-skin pixels such as background or covered body pixels~\cite{shaik2015comparative}. This technique has a wide range of applications, including human biometric analysis, medical image analysis, autonomous driving, and the beauty industry. Additionally, it is also important in other applications such as content retrieval, robotics, sign language recognition, and human tracking~\cite{mahmoodi2016comprehensive}. Furthermore, skin segmentation is often the first step in tasks related to nudity and appearance detection~\cite{maidhof2022underneath}. The use of cameras in video-based applications for Ambient Assisted Living (AAL) and remote monitoring of patients raises privacy concerns, particularly related to unwanted nudity. To address these concerns, a promising solution is privacy by context~\cite{padilla2015visual2}, which adapts privacy levels based on various factors, including the level of nudity. In these applications, the accurate detection of skin areas is critical for ensuring privacy.

However, skin segmentation is a challenging task as it faces many difficulties including variations in illumination conditions, different skin tones, camera variations, makeup, aging, and backgrounds with similar colors. The success of skin segmentation algorithms depends on their ability to overcome these challenges and accurately identify skin pixels in an image.

Prior to the advent of deep learning, most methods relied on the distribution of skin colors to identify fixed or adaptive ranges of skin pixel values in different color spaces or the training of a classifier according to each color space~\cite{he2019semi}. However, these methods have several drawbacks, such as poor generalization and high dependence on the training data. As a result, they tend to have poor performance in complex background situations and high false positive detection rates, making them ill-suited for real-world scenarios.

With the rise of deep learning, skin segmentation accuracy has improved drastically. The introduction of Fully Convolutional Networks (FCN) for semantic segmentation by Shelhamer et al.~\cite{shelhamer2016fully}, as well as increases in computation power, have boosted both accuracy and speed in the segmentation task. Additionally, other methods such as UNet~\cite{ronneberger2015u} and DeepLab~\cite{chen2017deeplab} were proposed, capable of segmenting multiple objects with high precision. These methods have also been adapted for human skin segmentation~\cite{he2019semi, zuo2017combining}. Despite the advances in accuracy achieved by these deep learning-based methods, there is still room for improvement. In particular, they are highly dependent on large datasets and tend to perform poorly on small or noisy skin segmentation datasets~\cite{hashemifard2022garment}.

In this paper, we propose a novel skin segmentation method that leverages the advances in human body segmentation to improve accuracy. In recent years, human body segmentation has improved significantly, with reliable open source models such as DensePose~\cite{guler2018densepose} and Mask R-CNN~\cite{he2017mask} being published, which can provide different body parts regions in a given image. These regions can serve as contextual information for detecting skin pixels. Instead of approaching the problem of skin segmentation from scratch, our method utilizes this auxiliary information to increase accuracy with lighter models. Our approach involves the use of two separate attention modules in the decoder, namely the Body Attention module and the Skin Attention module. The Body Attention module utilizes the output of the body mask to make the network focus on the body boundary. In the Skin Attention module, we make the assumption that the face and hands areas consist mostly of skin pixels. Therefore, we calculate the affinity of these pixels and body pixels embeddings to produce a skin attention mask, which provides extra guidance to the network for the target class. By integrating these attention modules, our method can effectively segment skin pixels by utilizing the information from human body segmentation and facial features.

Our contribution to the field of skin segmentation includes:
\begin{itemize}
\item A lightweight model that improves skin segmentation performance without greatly increasing model size.
\item Incorporating contextual information in the skin detection process to enhance performance.
\item Modifying commonly used attention mechanisms to better suit the skin segmentation task.
\item Utilizing individual skin color as a cue for segmentation to address variations in skin tone.
\item A weakly supervised training strategy that utilizes the proposed attention module for training on noisy datasets.
\end{itemize}

The above contributions provide a method that balances the trade-off between accuracy, resources and computational cost which make it feasible for real-world applications.

The remaining structure of this paper is as follows: Section 2 provides a brief overview of current methods for human skin detection. Section 3 describes the proposed method for skin segmentation, including the architecture and the details of attention modules. Section 4 presents experiments and results, as well as an ablation study. Section 5 introduces a weakly supervised training strategy to improve results. Finally, in Section 6, conclusions of the work and suggestions for future research are discussed.

\section{Related Work}
Human skin segmentation is a challenging task in computer vision and machine learning, and has been widely studied in the literature. Similar to many other recognition and segmentation tasks, human skin segmentation can be broadly categorized into two groups: traditional methods, and deep learning methods.

Traditional skin detection methods have primarily focused on skin color characteristics. For example, thresholding methods based on the intensity of skin color have been proposed in~\cite{chaves2010detecting, yang2002detecting}. These methods typically define a set of rules and conditions for each color channel, and check if a given pixel satisfies these conditions to decide if it belongs to the skin class or not. These rules are established either based on trial and error~\cite{gupta2016robust, vadakkepat2008multimodal, do2007skin} or by using thresholding algorithms~\cite{santos2016improved}. However, these methods have a high false detection rate and are strongly dependent on the training dataset and its conditions. 

To address the limitations of traditional skin detection methods, various researchers have proposed more robust methods by dynamically updating rules using cues in the image. For example, Shifa et al.~\cite{shifa2020skin} proposed a hybrid method with Combined Threshold-rules that adapts the threshold ranges by detecting the sampling skin tone and analyzing the color histogram and distribution. Probabilistic methods were also proposed as a way of detecting skin by evaluating the general distribution of skin color. These methods include using parametric and non-parametric techniques such as histograms, look-up tables (LUT), Naive Bayes and Gaussian distributions. For example, Gomez et al.~\cite{gomez2002selecting} used 3D histograms to propose the conditions for skin probability, utilizing a simplified version of probability theory. Nanni et al.~\cite{nanni2014effective} proposed the use of histograms with multiple LUTs in different color spaces, which resulted in a reduction of the false detection rate. However, these methods, despite their speed, do not take into account the relation between adjacent pixels and the spatial information which are crucial for skin detection~\cite{naji2019survey}. Bayesian classifiers have also been used for skin detection, but they require a large training set to achieve high accuracy. For instance,~\cite{nadian2016pixel, jones2002statistical, sigal2004skin} used the Bayes rule to calculate the conditional probability density function of a pixel from normalized histograms and adjust threshold values. Additionally, some works~\cite{caetano2002performance, liu2005efficient, shih2008extracting} have modeled skin color distribution using Gaussian mixtures, 2D Gaussian Probability Density Function (PDF) were used to model skin color in different color spaces like YUV, HSV and RGB. These methods typically require the use of expectation maximization (EM) for parameter optimization~\cite{moon1996expectation}. Supervised learning and binary classification methods, such as Support Vector Machines (SVMs) and Multi-Layer Perceptrons (MLPs), have also been employed in skin detection. These methods train a classifier on extracted color-texture features, such as Histogram of Oriented Gradients (HOG), Local Binary Patterns (LBP), FAST and ORB, derived from blobs of an image ~\cite{li2015pornographic, zhuo2016orb}. For example, Han et al.~\cite{han2009automatic} used SVM for binary pixel-wise classification. Another example is the combination of SVM and Gaussian Mixture Model (GMM) proposed in~\cite{zhu2004adaptive}.

Region-based skin detection is another approach that has been used in skin detection. These techniques aim to make use of spatial information and connections in an image to identify regions with similar features, without building a skin color model. Region growing is one such technique where the segmentation process starts with selected seed pixels. The neighboring pixels are compared to the seed pixel and if they have similar properties, they are added to the seed pixel's region, which then grows. This process is repeated for new adjacent pixels until no neighboring pixels satisfy the conditions and the growth of the region stops. Variations of this method have been proposed in~\cite{abdullah2008skin, chen2007region, mahmoodi2014boosting} where different measurement techniques have been proposed to measure the similarity between adjacent pixels for assigning them to a region, such as Euclidean distance, color distance map or probability measuring.

With the advent of deep learning, newer ideas have been proposed for human skin segmentation, many of which draw inspiration from the broader research area of semantic segmentation or a combination of various deep learning concepts. Early work treated the problem as a classification task by dividing an image into smaller patches and using deep networks to perform binary classification between skin and non-skin classes~\cite{kim2017convolutional, lei2016skin, dourado2019domain}. For example, Lei et al.~\cite{lei2016skin} proposed a patch-based skin segmentation method using stacked autoencoders to extract discriminative features from the blobs in an image, but this approach is not efficient in terms of time and resources required and it does not take into account relations between patches and contextual information.

Fully Convolutional Networks (FCNs)~\cite{shelhamer2016fully} have become the main approach for various segmentation tasks, including skin segmentation, due to their ability to reduce the number of parameters through an encoder-decoder architecture. The encoder extracts features and performs down-sampling, while the decoder up-samples the features to the original input size. For example, Kim et al.~\cite{kim2017convolutional} evaluated some of the prevalent FCN architectures for skin segmentation. Chang-Hsian et al.~\cite{ma2018human} used pre-trained ResNet50 and transfer learning. Roy et al.~\cite{roy2021robust} combined conditional adversarial training approach \cite{isola2017image} along with U-Net~\cite{ronneberger2015u}. Zuo et al.~\cite{zuo2017combining} used RNN layers as they claim that CNN are not sufficient to model the relationship between adjacent pixels. He et al.~\cite{he2019semi} proposed a semi-supervised method which takes advantage of other similar datasets and used dual-task fully convolutional network which shares the encoder and two separate decoders for detecting both body and skin parts in a U-Net shape auto-encoder. Arsalan et al.~\cite{arsalan2020or} proposed an end-to-end semantic segmentation network with an outer residual skip connection to transfer the edge information from early layers to end layers.

A successful technique in semantic segmentation is the use of self-attention modules for modeling long-range dependencies within an image. This technique was originally proposed in~\cite{vaswani2017attention} for machine translation but has since been widely used in various tasks such as~\cite{lin2017structured, shen2018disan}. In computer vision, self-attention modules were introduced in~\cite{zhang2019self} to extract global dependencies of inputs for better image generation. Since then, various kinds of self-attention mechanisms have been proposed, ranging from simple lightweight methods such as~\cite{woo2018cbam} to more complex methods like~\cite{fu2019dual} that capture dependencies in different dimensions.

\section{Method}
In this section, we present a novel method for efficient skin segmentation. We begin by describing the task-oriented attention modules that form a key component of our approach. Next, we outline the general architecture of our network, which is based on a fully convolutional network (FCN) with encoder-decoder architecture. Finally, we discuss the efficiency of our approach, specifically in terms of the reduction of parameters in the encoder.

\subsection{Attention mechanisms}
\label{lab:attention_mechanisms}
The attention modules are designed to help the network focus on specific areas of an image that are relevant to the task at hand, such as skin pixels in this case, without adding a large amount of redundant parameters. The two attention mechanisms used in this work are:
\begin{itemize}
\item Body Attention: This is used to emphasize the body area in a given image, as the skin pixels are only present within the body boundaries. This module is used in an earlier stage of the decoder, and its purpose is to guide the network to focus on the body area and extract body-related information from the encoded data.
\item Skin Attention: This module is designed to compare the embeddings of pre-defined skin areas, such as the face and hands, to all other body embeddings. The purpose of this module is to provide auxiliary guidance to refine the output. It is implemented in a later stage of the decoder with a higher resolution, which leads to finer boundaries and reduces noise.
\end{itemize}

By using these two attention mechanisms, the network is able to perform more efficiently by focusing on the given task and reach higher accuracy using fewer parameters. In the following, each module is explained in detail.

\subsubsection{Body Attention module}
The Body Attention module aims to focus feature extraction on the body area. It is based on the Convolutional Block Attention Module (CBAM)~\cite{woo2018cbam} method for adaptive feature refinement, but with slight modifications to make it better suited for the task of skin segmentation.

CBAM produces attention maps along the channel and spatial dimensions sequentially. The module starts by generating an attention map for the channel dimension by using a fully connected layer and a sigmoid activation function to weigh the importance of each channel. Then, it generates an attention map for the spatial dimension by using max-pooling and average-pooling to compute the importance of each spatial location. These two attention maps are then multiplied element-wise to produce a final attention map. This map is then used to weight the feature maps and refine the feature extraction.

This modified version of CBAM is used in the Body Attention module to emphasize the body area in a given image, and guide the network to focus on the body area and extract body-related information from the encoded data. This helps to improve the performance of the network by focusing on the most important areas of the image.

For a given block input tensor \( F \in  \mathbb{R}^{C\times H\times W} \), a 1D channel attention \( M_c \in  \mathbb{R}^{C\times 1\times 1} \) and a 2D spatial attention map \( M_s  \in  \mathbb{R}^{1\times H\times W} \) are inferred~\cite{woo2018cbam}. According to~\cite{zeiler2014visualizing} each channel of feature tensor acts as an object detector, therefore channel attention tries to find meaningful objects in an image.

For the channel attention sub-module we followed the CBAM approach described in detail by Woo et al.~\cite{woo2018cbam}. Specifically, we use the following steps:
\begin{enumerate}
\item The input tensor is max-pooled and average-pooled simultaneously and squeezed spatially.
\item The descriptor vectors are then fed to a shared network which is a multi-layer perceptron (MLP) with one hidden layer. The size of the hidden layer is reduced from the input by a ratio of $\mathcal{r}$.
\item The outputs of the shared MLP are then merged by summation and a sigmoid function is applied to the result.
\item Finally, this channel attention vector is broadcasted to the input tensor $F$. 
\end{enumerate}

The spatial attention sub-module focuses on the area where the desired object (skin) appears. It uses a modified version of the CBAM approach to compute the spatial attention. In CBAM, the spatial attention is computed by applying average-pooling and max-pooling operations along the channel axis and concatenating them to generate an efficient feature descriptor. This descriptor is then convolved by a standard convolution layer. In our approach, we make use of the knowledge that skin is present within the body boundaries. We extract a body mask by using a pre-trained network and concatenate it to the original concatenation before the convolution layer. This forces the attention module to put emphasis on the area of interest (human body) and extract more related attention maps from the input. The Sigmoid function is again applied to the final result. It has been empirically confirmed in our experiments that by adding the body mask in the concatenation operation, the network can extract a more related attention map from the input, which results in an improvement in accuracy of the whole network.

This refined and task-oriented CBAM sub-module can be summarized as follows:
\begin{equation}
     F^\prime = M_c(F) \otimes F
\end{equation}
\begin{equation}
     F^{\prime \prime} = M_s(F^\prime) \otimes F^\prime
\end{equation}

\noindent where:
\begin{equation}
     M_c(F) = \sigma(MLP(AvgPool(F)) + MLP(MaxPool(F)))
\end{equation}

\noindent and:
\begin{equation}
     M_s(F^\prime) = \sigma(Conv^{7\times 7}(Concat[AvgPool(F^\prime); MaxPool(F^\prime);Body Mask(I)]))
\end{equation}

\noindent The Sigmoid function is denoted by \( \sigma \), \(F^\prime \) is the input tensor to the spatial attention and \emph{I} is the input image. The output 2D map is multiplied to all input channels of \(F^\prime \). Finally, using a skip connection, the attention module output \(F^{\prime \prime} \) is summed with the module input tensor \(F \). 
%% The Appendices part is started with the command \appendix;
%% appendix sections are then done as normal sections
\subsubsection{Skin Attention Module}
Skin detection in images can be a difficult task, especially in unconstrained situations where skin areas can be exposed on different body parts with different shapes and under varying body poses. Conventional CNNs, which are designed to find objects and shapes for a given class, may not be completely effective or appropriate for this task. In addition, Fully Convolutional Networks with convolution operations have local receptive fields and fail to capture long-range contextual information. This means that the features extracted from the same class in different parts of the image may vary in representation, leading to a performance drop due to intra-class inconsistency. Furthermore, to account for the variety of human skin colors, the datasets for the deep learning model have to be large and inclusive of all skin types. To overcome these challenges, a skin attention module inspired by the work of Fu et al.~\cite{fu2019dual} is introduced in this study. The proposed method is able to take advantage of a wider range of contextual information and draw a global context over local features to improve feature representations for better pixel-level prediction and noise reduction. The mechanism is based on the assumption that face and hand areas of a given human image mostly consist of skin pixels.

In this proposed method, we begin by using a pre-trained body segmentation network to infer the body part masks from the input image, $I$. In the attention module,  an intermediate input feature tensor  \( T \in  \mathbb{R}^{C\times H\times W} \) is duplicated and then fed to a CNN layer followed by batch normalization to obtain \( {K,Q} \in  \mathbb{R}^{C\times H\times W} \). Then, by using the body part masks, the binary mask of face and hand area is multiplied by broadcasting to \emph{K}, and the full body binary mask is element-wise multiplied to \emph{Q}. Both outcomes are then reshaped to \( \mathbb{R}^{C\times N} \) where \( N = H \times W \). The energy matrix  \( E \in  \mathbb{R}^{N\times N} \) is calculated by taking the matrix multiplication of the transpose of \emph{Q} and \emph{K}.

By utilizing the body part masks, we then element-wise multiply the binary mask of the face and hand area with K and the full body binary mask with Q. Both matrices are reshaped to C$\times$N, where N=H$\times$W. The energy matrix, E, with dimensions N$\times$N, is calculated by taking the matrix multiplication of the transpose of Q and K.

Each non-zero column in the reshaped matrices \emph{K} and \emph{Q} represents the feature vector of face pixel and body pixel, respectively. Therefore, each element in \emph{E} represents the inner product or similarity measure between a pixel in the face and hands areas and a pixel in the body. By calculating the weighted average of \emph{E} with respect to the number of face and hands pixels in \emph{I}, the average similarity between the embedding of a pixel in body area and all the pixels in hands and face areas is obtained. The closer they are, the output of the inner product is higher. 

The resulting matrix, called the similarity matrix $S \in  \mathbb{R}^{N\times 1}$ is further processed by applying the tanh function for normalization and limiting the values below 1. It is then reshaped to \( 1 \times H \times W \), which has the same dimensions as the input tensor. This resulting matrix, called the attention map \emph{A}, illustrates the correlation between a body pixel and all supposed skin pixels in the hands and face of a person. Since the body of each person is compared to their corresponding face and hands, this method works well with a variety of skin colors. Additionally, all body embeddings are compared to the face regardless of location, body part, or the shape of that body part. This allows for a broader range of contextual information to be involved in the process, rather than relying solely on local features from the CNN.

To enable the network to adjust the effect of this attention map, a trainable weight parameter, \( \omega \), is applied to it. This allows the network to decide the importance of this attention map and use it accordingly. The final attention map is then added to the input tensor, \emph{T}, to obtain the final result.

\begin{figure}[t]
\centering
\includegraphics[width=.8\textwidth]{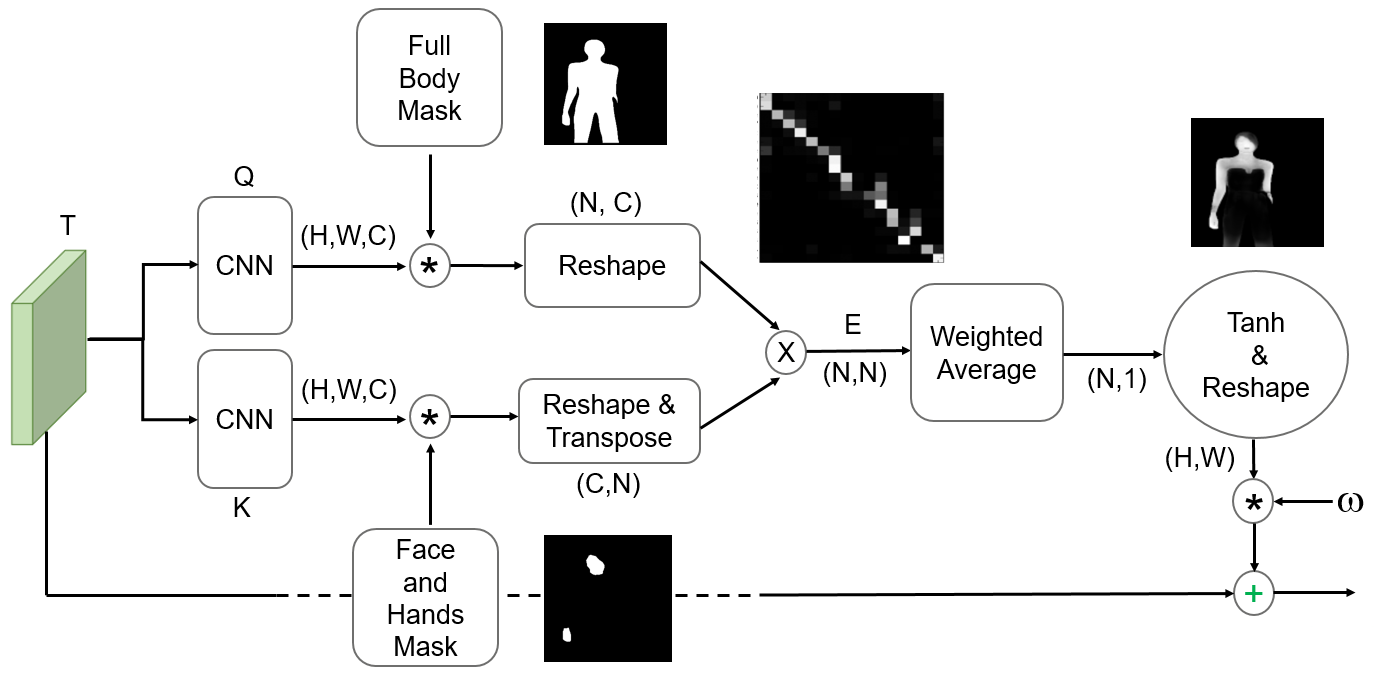}
\caption{Skin Attention Module. $\protect \ostar$ denotes element-wise multiplication and $\protect \times$ is matrix multiplication.}
\label{fig:pam module}
\end{figure}

Empirical results demonstrate that the trainable weight parameter, \( \omega \), tends to increase during the training process. This suggests that the attention map, \emph{A}, contains useful information for the final prediction. The proposed attention module, as a whole, is depicted in Figure~\ref{fig:pam module}. This highlights the effectiveness of the proposed method in capturing the correlation between pixels in the body, face and hands areas, and utilizing it to improve the final prediction.

\subsection{Network architecture}

Semantic segmentation has seen significant advancements in recent years, with several methods achieving high accuracy on benchmark datasets. However, these methods often rely on heavy backbone networks, making them less suitable for real-time applications with limited computation resources. To address this issue, researchers have proposed lightweight networks that can achieve performance comparable to that of high-quality networks while consuming less computational resources. Examples of such networks include ENet~\cite{paszke2016enet}, DFANet~\cite{li2019dfanet}, and LEDNet~\cite{wang2019lednet}, which employ techniques such as depthwise separable convolutions (DwConv2D)~\cite{chollet2017xception}, feature aggregation sub-networks, and channel split and shuffle.

The proposed model for semantic segmentation comprises of two channels. The primary channel of the network is an asymmetric encoder-decoder architecture that incorporates auxiliary sub-modules in the pipeline. It is inspired by HLNet~\cite{feng2020hlnet} and HRNet~\cite{sun2019deep} and employs best practices in segmentation modules.

The encoder consists of CNNs and DwConv2D, bottlenecks and interaction modules. The input image, with a resolution of \( 256\times 256 \), is fed to a CNN block with 32 filters, followed by two DwConv2D blocks with 64 filters and a stride of 2. This fast downsampling process ensures low-level feature sharing~\cite{poudel2019fast}. Each block has \( 3\times 3 \) kernels, followed by batch normalization and ReLU activation. To preserve details and constrain the number of parameters, the maximum downsampling rate is set to 1/8. 
The output, with a resolution of \( 32\times 32 \), is then fed to the information interaction module proposed in~\cite{feng2020hlnet}. This module consists of three parallel inverted residual blocks using different filter sizes and strides. Each block captures information with different resolutions and feature map sizes, learning multi-scale information representation. This process increases the network's ability to segment small objects and shapes while preserving more details.
Subsequently, the information from high to low resolutions is combined together by concatenation. For more details please refer to HLNet~\cite{feng2020hlnet}.

In addition to the primary channel, the proposed model also includes an auxiliary channel that is pre-trained for body segmentation. Given an image, this model produces a whole body mask, as well as separate masks for the face and hands. These masks are then resized and fed to the decoder to be used in attention mechanisms. These attention mechanisms (see Section~\ref{lab:attention_mechanisms}) aim to make use of the information provided by the auxiliary channel to improve the overall performance of the semantic segmentation.

The decoder of the proposed model consists of three bilinear upsampling layers, each followed by a convolution block to construct an original size image from the \( 32\times 32 \) feature map in the bottleneck. To make use of the information provided by the auxiliary channel, a body attention module is applied after the first upsampling layer and a skin attention module is applied after the second one. Additionally, skip connections are added to propagate the error to the early feature extraction layers.

\begin{figure}[t]
\centering
\includegraphics[width=.8\textwidth]{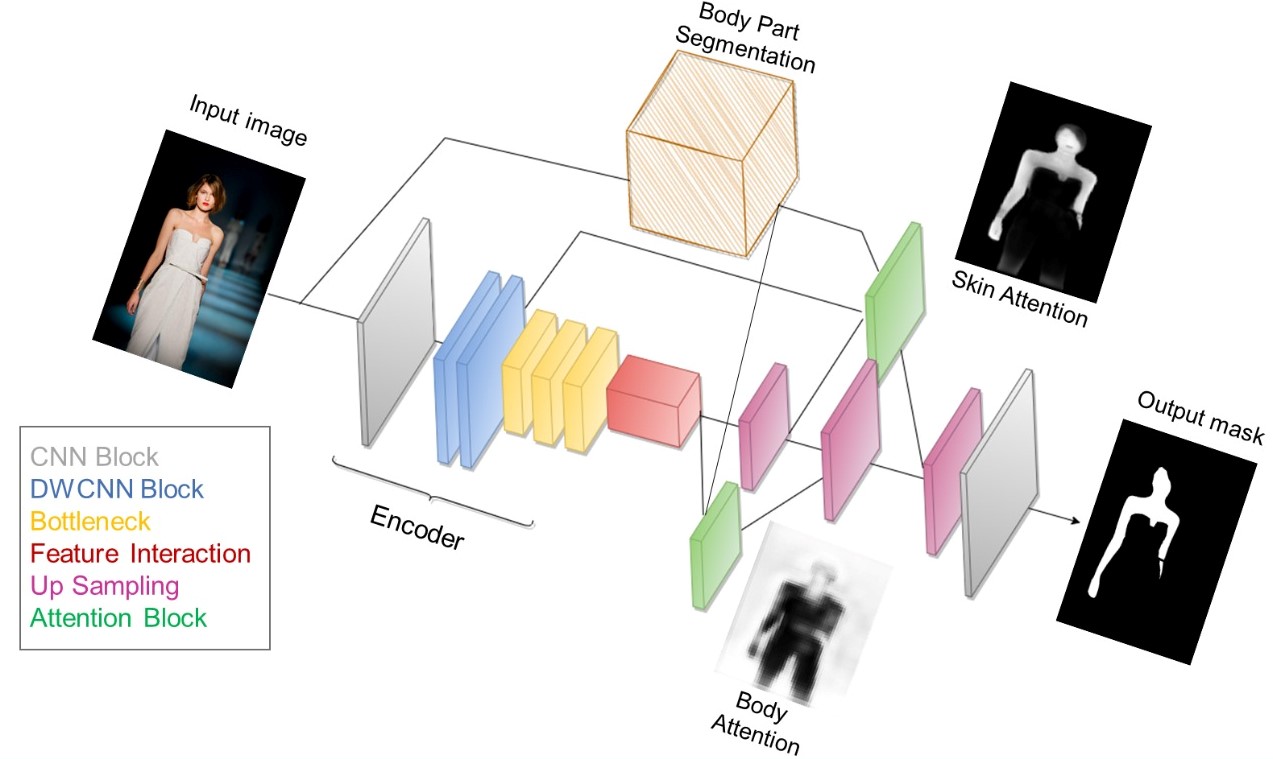}
\caption{Proposed skin segmentation network}
\label{fig:network}
\end{figure}

Finally, a Sigmoid layer is applied to perform binary classification between skin and non-skin pixels. In order to optimize the network, given the class imbalance between skin and non-skin pixels in most images, where the background is larger than the area covered by the person, a combination of Dice loss and Binary Focal loss is used between the output mask and the ground truth. This has been shown to be more effective for segmenting small objects~\cite{furtado2021testing}. The overall network architecture is illustrated in Figure~\ref{fig:network}.

\section{Experimental Results}
\subsection{Datasets and implementation details}
Obtaining data for skin segmentation is a common challenge in this field. One approach is to collect bespoke datasets, but there are also several public datasets available for this task which can be useful for the evaluation and comparison of methods, even though they may not follow the same protocol (i.e., some considered eyebrows and lips as skin and some excluded them) or contain noise. To train the proposed method, the visuAAL Skin Segmentation dataset~\cite{hashemifard2022garment} (VSS) is used. The VSS dataset contains 46,775 high-quality images divided into a training set with 45,623 images and a validation set with 1,152 images. The skin labels were automatically extracted with an algorithm. Additionally, 230 images were manually segmented for evaluation purposes, which are used as the test set in order to report the performance. Another dataset incorporated in this work is the Pratheepan face Dataset~\cite{tan2011fusion}. Pratheepan is a small dataset containing 78 images with precise annotation and is mostly used as a benchmark. The obtained results are evaluated on this dataset in order to make a comparison with existing methods. Both datasets are publicly available.

In this work, the DensePose method was used for body, face and hands area detection. DensePose is a method that can provide precise body masks. To incorporate this information in our  network, the parameters of the DensePose network were frozen during the training process.

The proposed method was implemented using TensorFlow and the training was run on two GPUs, one NVIDIA GeForce GTX 1080Ti and one NVIDIA GeForce GTX 2080Ti. The network was trained for 30 epochs using an initial learning rate of 0.001, with a decay rate of 0.96, and the ADAM optimization algorithm. The whole training process took about 5 hours.

\subsection{Results}

The results of the proposed method were compared with other important semantic segmentation methods on the VSS dataset in Table~\ref{tab:result}. All of these networks were trained and evaluated on the same data. As shown in the table, the proposed method outperformed all other methods with a considerably lower number of parameters. One important thing to note is the improvement in results when comparing HLNet to our method. Despite their similarities in terms of the backbone network architecture and the amount of parameters employed, the proposed method shows significant improvement in all metrics for skin segmentation. This demonstrates the effectiveness of the proposed attention mechanisms.

\begin{table}[t]
% \small
    \centering
    \caption{Semantic Segmentation results on VSS dataset.}
    \label{tab:result}
    \begin{tabular}{lcccccc} \toprule
        \multicolumn{1}{p{1.5cm}}{ Method} & \multicolumn{1}{p{1cm}}{\centering Precision} & \multicolumn{1}{p{1cm}}{\centering Recall} & \multicolumn{1}{p{1cm}}{\centering F1-score} & \multicolumn{1}{p{1cm}}{\centering CDR} & \multicolumn{1}{p{1cm}}{\centering DSC} & \multicolumn{1}{p{1cm}}{\centering Parameters}\\ \midrule
        SegNet~\cite{badrinarayanan2017segnet} & 80.71\% & 80.12\% & 80.29\% & 97.75\% & 78.82\% &30M \\ 
        UNet~\cite{ronneberger2015u} & 82.66\% & 85.34\% &83.83\% & 98.01\% & 82.38\%  &30M\\ 
        Basic FCN~\cite{long2015fully} & 70.34\% & 84.88\% & 76.80\% & 97.11\% & 74.40\% & 10M \\ 
        DSNet~\cite{hasan2020dsnet} & 85.80\% & 85.08\% & 85.40\% & 98.35\% & 84.14\% &8M\\  
        HLNet~\cite{feng2020hlnet} & 76.50\% & 79.86\% & 78.01\% & 97.43\% & 76.08\% &1.2M\\
        \midrule
        \bf{Proposed Method} & \bf{88.30\%} &	\bf{85.91\%} & \bf{86.96\%} & \bf{98.45\%} &	\bf{86.44\%} & 1M \\
        Base Network & 76.20\% & 78.22\% & 77.19\% & 97.24\% & 75.88\% &1M\\ 
        With Body Attention & 78.11\% & 82.08\% & 80.04\% & 97.81\% & 78.92\% &1M\\  
        With Skin Attention& 86.62\% & 83.11\% & 84.82\% & 98.18\% & 83.33\% &1M\\ 
        \bottomrule
    \end{tabular}

\end{table}

\subsection{Ablation study}

\begin{figure}[t]
\centering
\begin{subfigure}[h]{0.2\textwidth}
    \centering
    \includegraphics[width=0.75\textwidth]{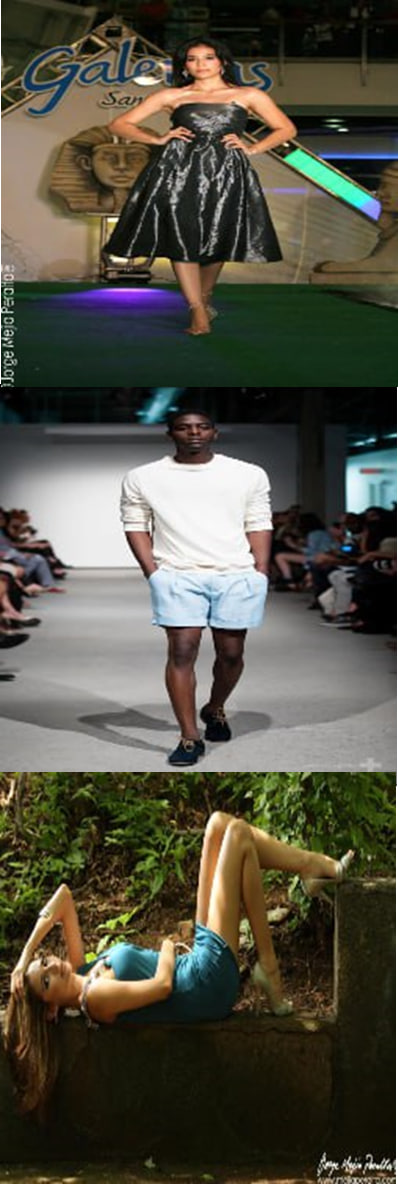}
    \caption{Original image}
    \label{fig:11}
\end{subfigure}
\hspace{.1em}
\begin{subfigure}[h]{0.2\textwidth}
    \centering
    \includegraphics[width=0.75\textwidth]{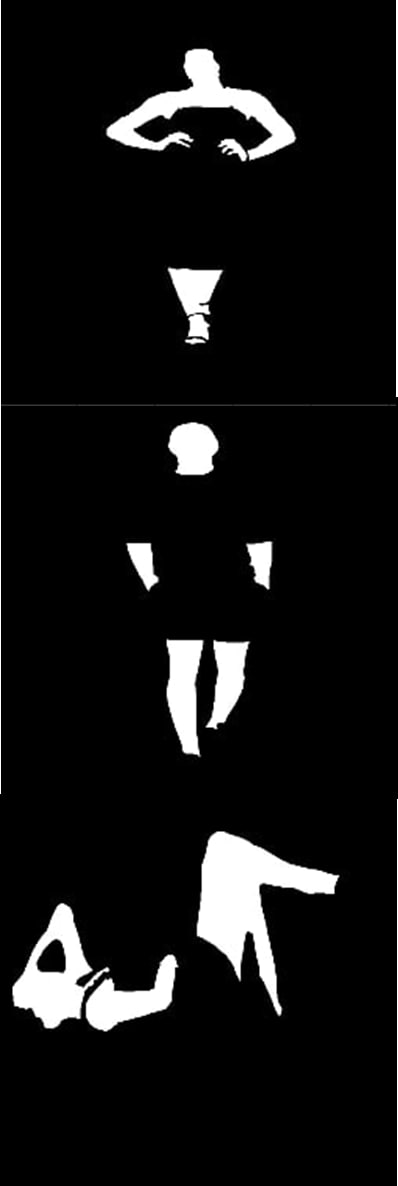}
    \caption{Ground truth}
    \label{fig:22}
\end{subfigure}
\hspace{.1em}
\begin{subfigure}[h]{0.2\textwidth}
    \centering
    \includegraphics[width=0.75\textwidth]{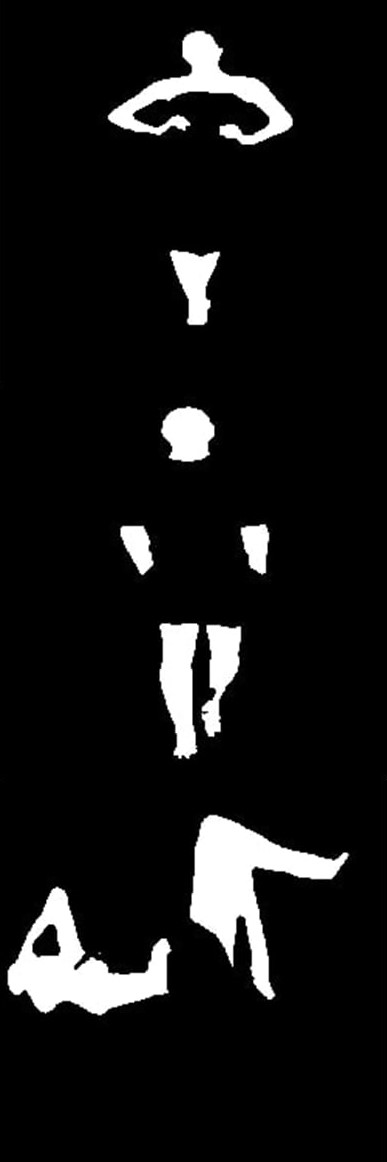}
    \caption{Model output}
    \label{fig:33}
\end{subfigure}
\hspace{.1em}
\begin{subfigure}[h]{0.2\textwidth}
    \centering
    \includegraphics[width=0.75\textwidth]{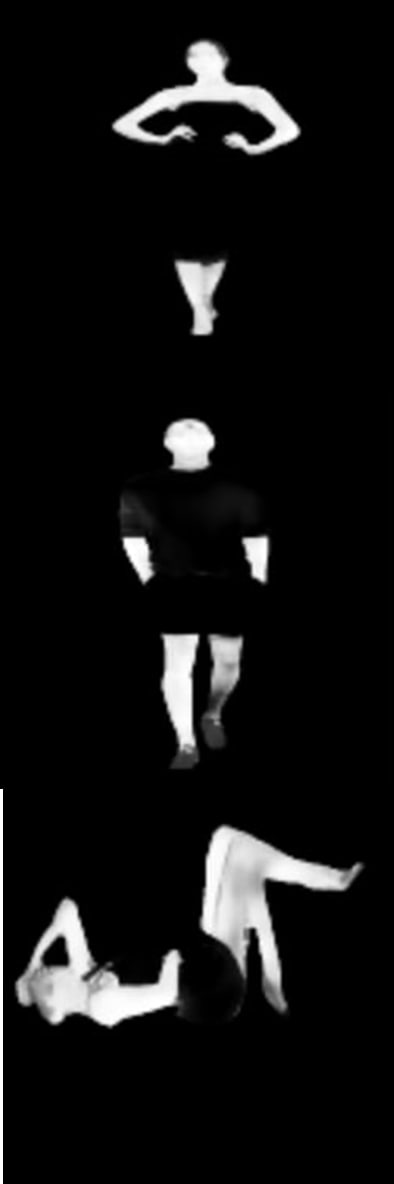}
    \caption{Attention map}
    \label{fig:44}
\end{subfigure}  
\hspace{.1em}

\caption{The proposed method segmentation results on VSS dataset}
\label{fig:vss}
\end{figure}

Furthermore, to validate the effect of each module, each of them was evaluated separately. The results are presented in Table~\ref{tab:result}, and it is shown that the Skin Attention played a more important role compared to the Body Attention module, yet adding Body Attention still improved the results from the base network. Figure~\ref{fig:vss} illustrates some of the segmentation results. One significant observation in the proposed method is the impact of the Body Attention and Skin Attention modules on Recall and Precision. The Body Attention module has a greater effect on Recall compared to Precision, while the Skin Attention improves Precision to a greater extent than Recall. This can be explained by the Body Attention's emphasis on the whole body area, resulting in the inclusion of all body pixels and thus, a higher Recall. Conversely, the Skin Attention focuses on excluding non-skin pixels from the final mask, leading to improved Precision.
Furthermore, the \( \omega \) parameter which adjusts the amount of effect and participation of the Attention map in the whole network is set to 1.3 at the end of the training process.

\section{Improvement using weakly supervised training for noise reduction}

Deep learning methods have shown better performance compared to previous approaches for semantic segmentation, but they require a large amount of training data to perform adequately. The annotation process for segmentation is costly, very demanding and labor-intensive. Most of the datasets for skin segmentation are either small or suffering from low-quality images, but the main problem is the labelling noise as it is expensive to annotate such a large number of images accurately.

To make use of these big but noisy datasets, a method is proposed to modify the ground truth labels during a recursive training process. This is addressed as a weak supervision task in which, although the labels for a desired class exist (skin pixels), they may be annotated wrongly. In other words, every pixel with a skin class label can either belong to skin or non-skin class, but the pixels labelled as non-skin are considered to be correct. This assumption is made after studying available skin datasets. In this work, the VSS dataset is used, which includes a large training set produced automatically and may contain some noise, and a small fully supervised validation set. The problem of training directly over this dataset is that for a number of epochs, the improvement over validation and training sets accuracy can be seen. However, as the number of epochs increases, the network tends to overfit on the noisy training data, and this causes a drop in validation accuracy. To address this problem, the training data is utilized to the point that it is useful for the objective, and then the noise is modified to redirect the network parameters to segment the desired areas.

This approach allows the network to make use of the large amount of training data while addressing the problem of noisy labels in the dataset. The modified training set is then used to continue training the network for a fixed number of rounds. With this method, the network learns the general segmentation task in the warm-up step and then is able to improve its performance by using the Attention map to correct the noisy labels in the training set in each modification step. The threshold used to relabel the pixels is increased after each modification step, allowing the network to focus on pixels that are more likely to belong to the skin class while ignoring the noisy labels. Additionally, pixels outside the body boundary area are not relabelled, providing an additional constraint on the modification process.

The Attention map is used before applying the weight parameter \( \omega \) in the skin attention module. As the Attention map is computed as the product of the tanh function, the values are limited to 1, and can be considered as a skin probability map regarding face and hand pixels. This probability map is then multiplied with the weight parameter \( \omega \) to adjust the effect and participation of the Attention map in the overall network.

The approach of using the Attention map to modify the ground truth labels during the training process is inspired by previous work~\cite{khoreva2017simple, germi2022enhanced}. However, two modifications have been made. First, instead of using the output after each round, the Attention map is used. This provides stronger supervision as it uses the auxiliary information provided by the body segmentation module. Second, instead of searching in the rectangular bounding box, the search is limited to the body area (i.e., the only place skin pixels can exist). This reduces the search space and increases the precision of the relabelling process. Additionally, the spatial continuity condition is skipped in this approach. This is because, depending on the garment types, skin exposure can be found in any shape and size over the body.

The labelling procedures for a given pixel in position \( (i,j) \) can be written as:
\begin{equation}
  L_{(i,j)}^{new} = 
    \begin{cases}
      1 & \text{if }  L_{(i,j)}^{prev} \otimes P(S|F,H)_{(i,j)} > t \\
      0 & \text{otherwise}
    \end{cases}       
\end{equation}

\noindent in which \( L^{new} \) will be the new label after this training  round, \( L^{prev} \) is the last round training ground truths, \( P(S|F,H) \) is the probability map of the skin class given hand and face and \( t \) is the threshold value.

\begin{figure}[t]
\centering

\begin{subfigure}[t]{0.12\textwidth}
    \includegraphics[width=\textwidth]{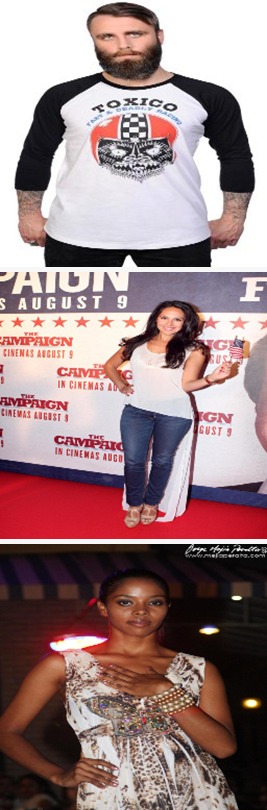}
    \caption{}
    \label{fig:first}
\end{subfigure}
\hspace{.1em}
\begin{subfigure}[t]{0.12\textwidth}
    \includegraphics[width=\textwidth]{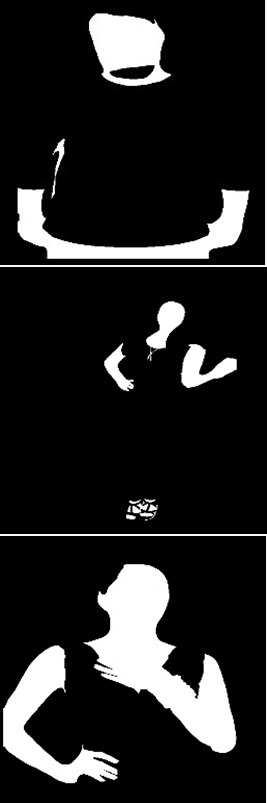}
    \caption{}
    \label{fig:second}
\end{subfigure}
\hspace{.1em}
\begin{subfigure}[t]{0.12\textwidth}
    \includegraphics[width=\textwidth]{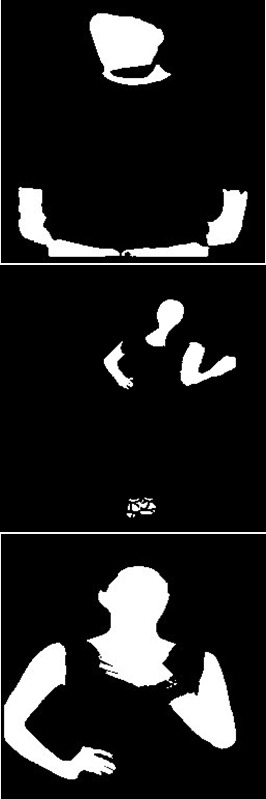}
    \caption{}
    \label{fig:third}
\end{subfigure}
\hspace{.1em}
\begin{subfigure}[t]{0.12\textwidth}
    \includegraphics[width=\textwidth]{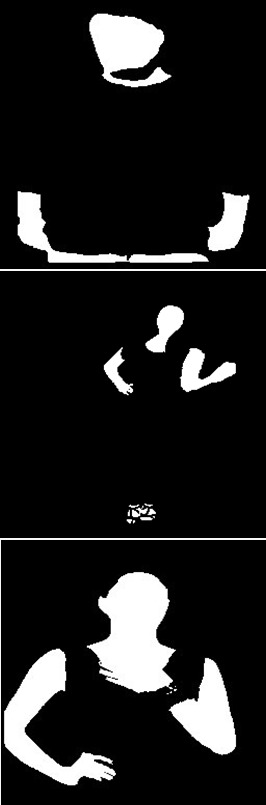}
    \caption{}
    \label{fig:fourh}
\end{subfigure}  
\hspace{.1em}
\begin{subfigure}[t]{0.12\textwidth}
    \includegraphics[width=\textwidth]{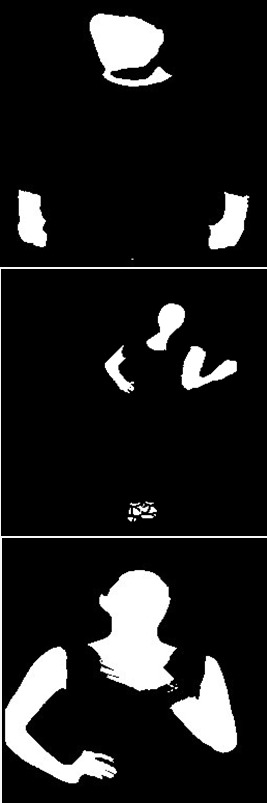}
    \caption{}
    \label{fig:fitht}
\end{subfigure} 

\caption{Recursive training effect on the noisy labels. (a) Original image, (b) noisy labels, (c) after 1st modification round, (d) after 2nd modification round, (e) after 4th modification round}
\label{fig:mask_errors}
\end{figure}

The improvement in performance is measured using the validation set after each round of the modification step. This process is continued until a performance drop is observed in the validation set. This indicates that the threshold for relabelling pixels has been increased so much that skin pixels are being discarded from the new ground truth labels. It is important to note that this procedure may not remove the noise from all images in the dataset equally or to the same extent, but it corrects the problematic and faulty labels that cause the most errors. As a result, the overall quality of the segmentation improves. In this way, it makes the performance higher on the denoised validation data. For the final step of training, the ground truth labels produced in the epoch previous to the performance drop are considered as the final training set. The training is continued over these labels until the best results are achieved. Depending on the level of noise in each image, the number of modification epochs needed to denoise might be different. As illustrated in Figure~\ref{fig:mask_errors}, for a small modification such as removing a bracelet or necklace, even one epoch is enough, however, large mislabeling requires more rounds of this recursive training procedure.

\begin{table}[t]
    \centering
    \caption{Comparing the results of the proposed method on the VSS dataset before and after recursive training.}
    \label{tab:result2}
    \begin{tabular}{lccccc} \toprule
        \multicolumn{1}{p{1.5cm}}{ Method} & \multicolumn{1}{p{1cm}}{\centering Precision} & \multicolumn{1}{p{1cm}}{\centering Recall} & \multicolumn{1}{p{1cm}}{\centering F1-score} & \multicolumn{1}{p{1cm}}{\centering CDR} & \multicolumn{1}{p{1cm}}{\centering DSC} \\ \midrule

        Base Network & 76.20\% & 78.22\% & 77.19\% & 97.24\% & 75.88\% \\ 
        Proposed Method & 88.30\% & 85.91\% & 86.96\% & 98.45\% & 86.44\% \\  
        \bf{Recursive Training} & \bf{90.59\%} &	\bf{87.92\%} & \bf{89.17\%} & \bf{98.61\%} &	\bf{88.91\%}  \\
        \bottomrule
    \end{tabular}
\end{table}

In this experiment, the number of training rounds \( n \) used for the warm-up step was 2, and the threshold value \( t \) for modification started at 0.2 with an increasing step size of 0.05 after each round. The optimum threshold reached at the end was 0.35 and the normal training process started with the dataset generated by this round. In Table~\ref{tab:result2}, the improvement of the results compared with the direct training approach is shown. It can be seen that the proposed method outperforms the direct training approach in terms of overall segmentation performance. This demonstrates the effectiveness of the proposed method in addressing the problem of labelling noise in the training dataset and improving the segmentation results.

\begin{figure}[t]
\centering
\begin{subfigure}[h]{0.2\textwidth}
    \centering
    \includegraphics[width=0.75\textwidth]{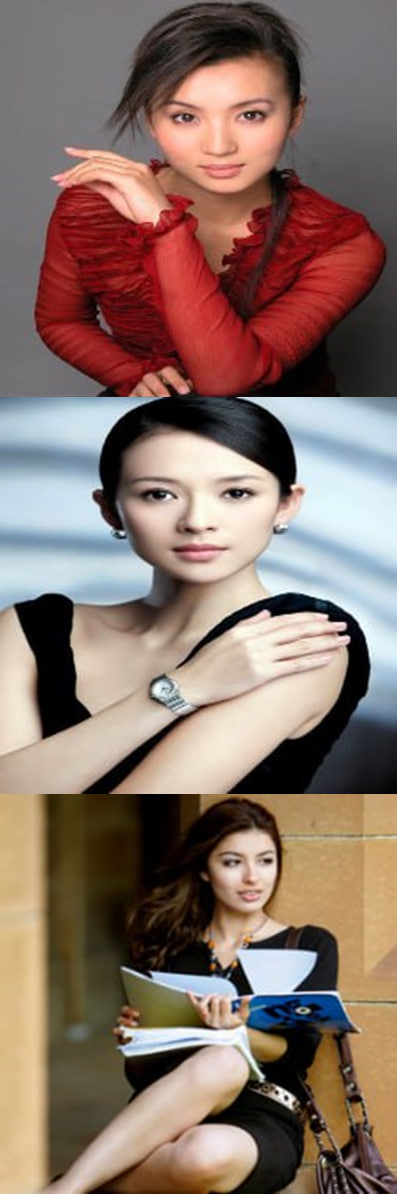}
    \caption{Original image}
    \label{fig:1}
\end{subfigure}
\hspace{.1em}
\begin{subfigure}[h]{0.2\textwidth}
    \centering
    \includegraphics[width=0.75\textwidth]{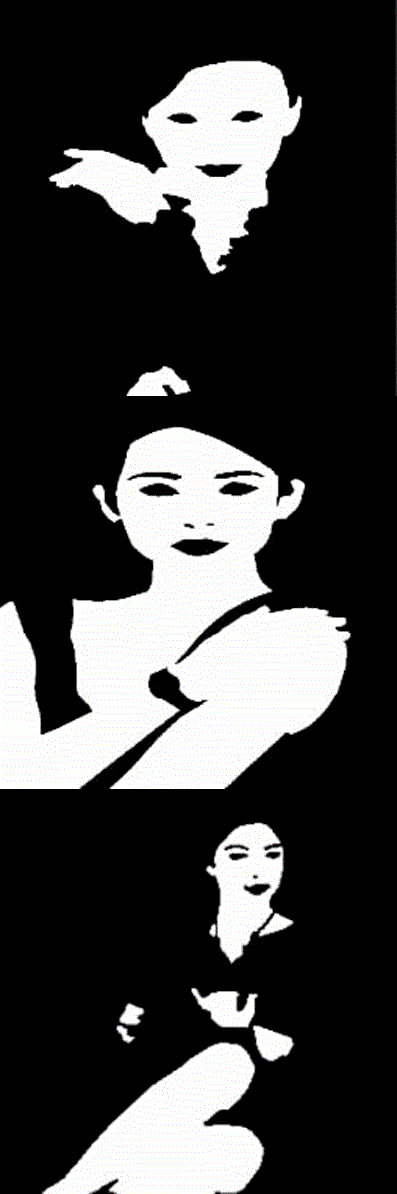}
    \caption{Ground truth}
    \label{fig:2}
\end{subfigure}
\hspace{.1em}
\begin{subfigure}[h]{0.2\textwidth}
    \centering
    \includegraphics[width=0.75\textwidth]{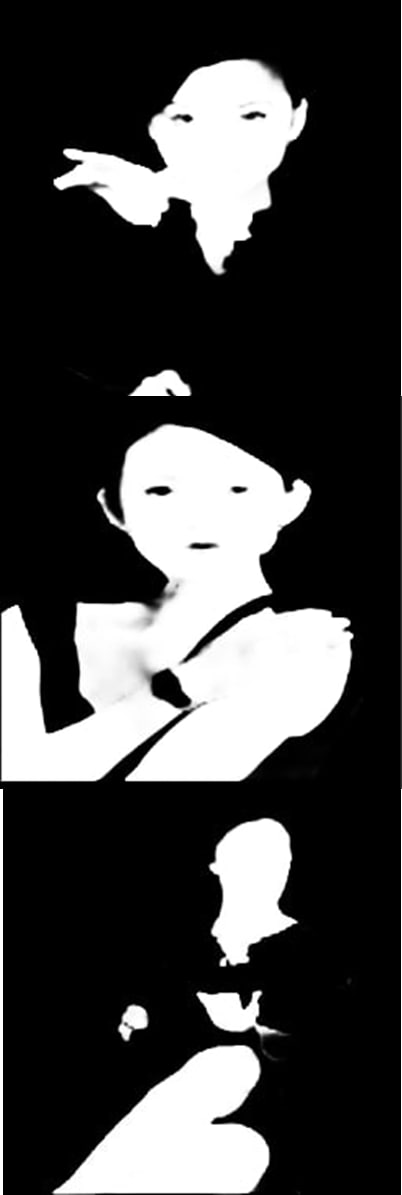}
    \caption{Model output}
    \label{fig:3}
\end{subfigure}
\hspace{.1em}

\caption{Segmentation results on the Pratheepan benchmark dataset}
\label{fig:pratheepan}
\end{figure}

In addition, in order to make a comparison between our method and other skin segmentation methods, the evaluation, over the Pratheepan dataset, of the proposed method after the recursive training steps was performed and is illustrated in Table \ref{tab:result3}. As it can be seen, our method is performing better than classic approaches and produces very similar results to the state-of-the-art method. Considering the model sizes, this can be considered as a higher improvement in efficiency. Some samples of the results on the Pratheepan dataset are illustrated in Figure \ref{fig:pratheepan}. This comparison further validates the effectiveness of the proposed method in addressing the problem of labelling noise and improving the overall segmentation performance while also being more efficient than other methods.

\begin{table}[t]
    \centering
    \caption{Evaluation of human skin segmentation methods on the Pratheepan dataset.}
    \label{tab:result3}
    \begin{tabular}{lccc} \toprule
        \multicolumn{1}{p{4cm}}{ Method} & \multicolumn{1}{p{1.5cm}}{\centering Precision} & \multicolumn{1}{p{1.5cm}}{\centering Recall} & \multicolumn{1}{p{1.5cm}}{\centering IoU} \\ \midrule
        
        Thresholding~\cite{kovac2003human} & 65.31\% & 89.58\% & 60.20\% \\ 
        GMM~\cite{jones2002statistical} & 62.36\% & \bf{91.5\%} & 60.46\% \\ 
        UNet~\cite{ronneberger2015u} & 89.55\% & 87.87\% & 85.50\% \\ 
        SOTA~\cite{he2019semi} & \bf{92.58\%} & 87.51\% & 87.90\% \\  
        \bf{Our Method} & 89.24\% &	90.95\% & \bf{89.80\%}  \\
        \bottomrule
    \end{tabular}

\end{table}

\section{Conclusion}
In conclusion, a lightweight, efficient and robust model for human skin segmentation is proposed in this paper. By utilizing prior knowledge and contextual information, the proposed method addresses some of the main challenges in human skin detection, such as variations in skin color and real-time performance. Additionally, a weakly supervised training strategy is proposed using the attention module to make large datasets with possible annotation errors more usable. The results show that the proposed method outperforms other existing methods in terms of accuracy and efficiency. Furthermore, the method is able to handle unseen skin characteristics and colors. However, to further improve the model, future enhancements such as reducing the memory requirements for calculating the skin attention, adding post-processing steps for smoothing the detected regions and refining the output, and keeping consistency may be beneficial.

\section*{Acknowledgments}
This work is part of the visuAAL project on Privacy-Aware and Acceptable Video-Based Technologies and Services for Active and Assisted Living~(\url{https://www.visuaal-itn.eu/}). This project has received funding from the European Union’s Horizon 2020 research and innovation programme under the Marie Skłodowska-Curie grant agreement No 861091.

%Bibliography
\bibliographystyle{unsrt}  
\bibliography{references}

\end{document}